\documentclass[a4paper, 5p]{elsarticle}

\usepackage{lineno}
\usepackage{amssymb}
\usepackage{graphicx}
\usepackage{hyperref}
\usepackage{url}

\biboptions{super}

\journal{Elsevier}

\begin{document}

\begin{frontmatter}

\title{Understanding electricity prices beyond the merit order principle using explainable AI}

\author[ste,col]{Julius Trebbien\corref{cor}}
\ead{j.trebbien@fz-juelich.de}
\author[NMBU,OsloMet]{Leonardo Rydin Gorj\~ao}
\author[RWTH,jara]{Aaron Praktiknjo}
\author[kit]{Benjamin Sch\"afer} 
\author[ste,col]{Dirk Witthaut}
\ead{d.witthaut@fz-juelich.de}

\address[ste]{Forschungszentrum J\"ulich, Institute for Energy and Climate Research -- Systems Analysis and Technology Evaluation (IEK-STE), 52428 J\"ulich, Germany}
\address[col]{University of Cologne, Institute for Theoretical Physics, Z\"ulpicher Str.~77, 50937 Cologne, Germany}
\address[NMBU]{Faculty of Science and Technology, Norwegian University of Life Sciences, 1432 As, Norway}
\address[OsloMet]{Department of Computer Science, OsloMet -- Oslo Metropolitan University, 0130 Oslo, Norway}
\address[RWTH]{Chair for Energy System Economics, Institute for Future Energy Consumer Needs and Behavior (FCN), E.ON Energy Research Center, RWTH Aachen University, Mathieustr.~10, 52074 Aachen, German}
\address[kit]{Institute for Automation and Applied Informatics, Karlsruhe Institute of Technology, 76344 Eggenstein-Leopoldshafen, Germany}
\address[jara]{JARA-ENERGY, 52074 Aachen, Germany}
            
\cortext[cor]{Corresponding author}

\date{\today}

\begin{abstract} 
Electricity prices in liberalized markets are determined by the supply and demand for electric power, which are in turn driven by various external influences that vary strongly in time. In perfect competition, the merit order principle describes that dispatchable power plants enter the market in the order of their marginal costs to meet the residual load, i.e.~the difference of load and renewable generation.
Many market models implement this principle to predict electricity prices but typically require certain assumptions and simplifications.
In this article, we present an explainable machine learning model for the prices on the German day-ahead market, which substantially outperforms a benchmark model based on the merit order principle. 
Our model is designed for the ex-post analysis of prices and thus builds on various external features. Using Shapley Additive exPlanation (SHAP) values, we can disentangle the role of the different features and quantify their importance from empiric data. 
Load, wind and solar generation are most important, as expected, but wind power appears to affect prices stronger than solar power does. 
Fuel prices also rank highly and show nontrivial dependencies, including strong interactions with other features revealed by a SHAP interaction analysis. Large generation ramps are correlated with high prices, again with strong feature interactions, due to the limited flexibility of nuclear and lignite plants. 
Our results further contribute to model development by providing quantitative insights directly from data.
\end{abstract}


\end{frontmatter}

\section*{Highlights}

\begin{itemize}
\item Explainable ML model outperforms benchmark model based on the merit order principle.
\item SHAP analysis reveals which features affect electricity prices beyond residual load.
\item Load, wind and solar generation are key features, but dependencies differ slightly.
\item Model quantifies the impact of fuel prices as well as generation ramps.
\end{itemize}



\section{Introduction}
\label{sec:intro}

The reliable supply of electric power is vital for modern societies \cite{van_der_vleuten_transnational_2010,praktiknjo2016value}. A stable operation of the electric power system requires that power generation and load are always balanced \cite{wood2013power}. Electricity markets are central to coordinate generation and demand; from long-term future contract to short-term spot trading. This coordination is becoming increasingly challenging due to the ongoing energy transition \cite{witthaut2022collective, milano_foundations_2018}. Generation from renewable sources such as wind and solar power is determined by the weather and thus highly volatile \cite{staffell2018increasing}, making also the electricity price highly volatile \cite{han_complexity_2022}.

European electricity markets were liberalized starting in the 1990s \cite{jamasb2005electricity}. 
Before, generation, transmission, and distribution were typically integrated in a single company holding a regional monopoly. 
In liberalized markets, different suppliers compete which should lead to an improved efficiency and lower costs \cite{jamasb2005electricity}.
Today, several markets exist in Europe which enable electricity trading on different time horizons \cite{epex2019annual}. 
In particular, the day-ahead markets of the European Power Exchange (EPEX SPOT) cover 13 countries and reached a market volume of more than 500 TWh in 2019 \cite{epex2019annual}. 
Recently, European energy markets were heavily disturbed by the Russian invasion of Ukraine causing an intensified research interest (see, e.g.~\cite{osivcka2022european,zakeri2022energy}).

The functioning and design of electricity markets is a central topic in energy economics \cite{stoft2002power,bublitz_survey_2019}. Mechanistic models, such as agent-based models, start from first principles and derive statements that can be verified or falsified on data \cite{hansen2019agent,reeg_amiris_2012}. Data-based approaches act complementary, i.e., they start from data and make no assumptions of underlying mechanisms. General rules that govern the system are inferred, providing valuable inputs to improve mechanistic models.
In recent years, machine learning has become an important method in electricity market research. The vast majority of studies focuses on the forecasting of electricity prices \cite{lago_forecasting_2021}. However, modern developments in interpretable machine learning enable much richer applications. In particular, they allow for in-depth scientific insights from large heterogeneous data sets \cite{roscher2020explainable}.

In this article, we establish a machine learning model for the electricity prices on the German day-ahead spot market. We apply SHapley Additive exPlanations (SHAP) \cite{lundberg_local_2020} to explain the model and provide insights into which factors determine the market price. We use an extended data set to identify driving factors which are commonly neglected in elementary studies or machine learning models \cite{lago_forecasting_2021}. The model substantially outperforms a benchmark model based on a common merit order principle and thus reveals additional details into the function of the market. For instance, the model quantifies the impact of fossil fuel prices and load ramps, as well as nonlinear interactions of different features. 

The remaining article is structured as follows. In Sec.~\ref{sec:background}, we provide an introduction to the German electricity market, market models based on the merit order principle and review previous works in the field. In Sec.~\ref{sec:methods}, we discuss how we obtained and processed data, how the machine learning model is trained and interpreted. We then continue in Sec.~\ref{sec:results} to analyse the results of the machine learning model, in particular demonstrating how load, wind and solar generation but also fuel prices critically influences electricity prices. We close in Sec.~\ref{sec:discussion} with a discussion.

\begin{figure*}[tb]
    \centering
    \includegraphics[width=0.9\textwidth]{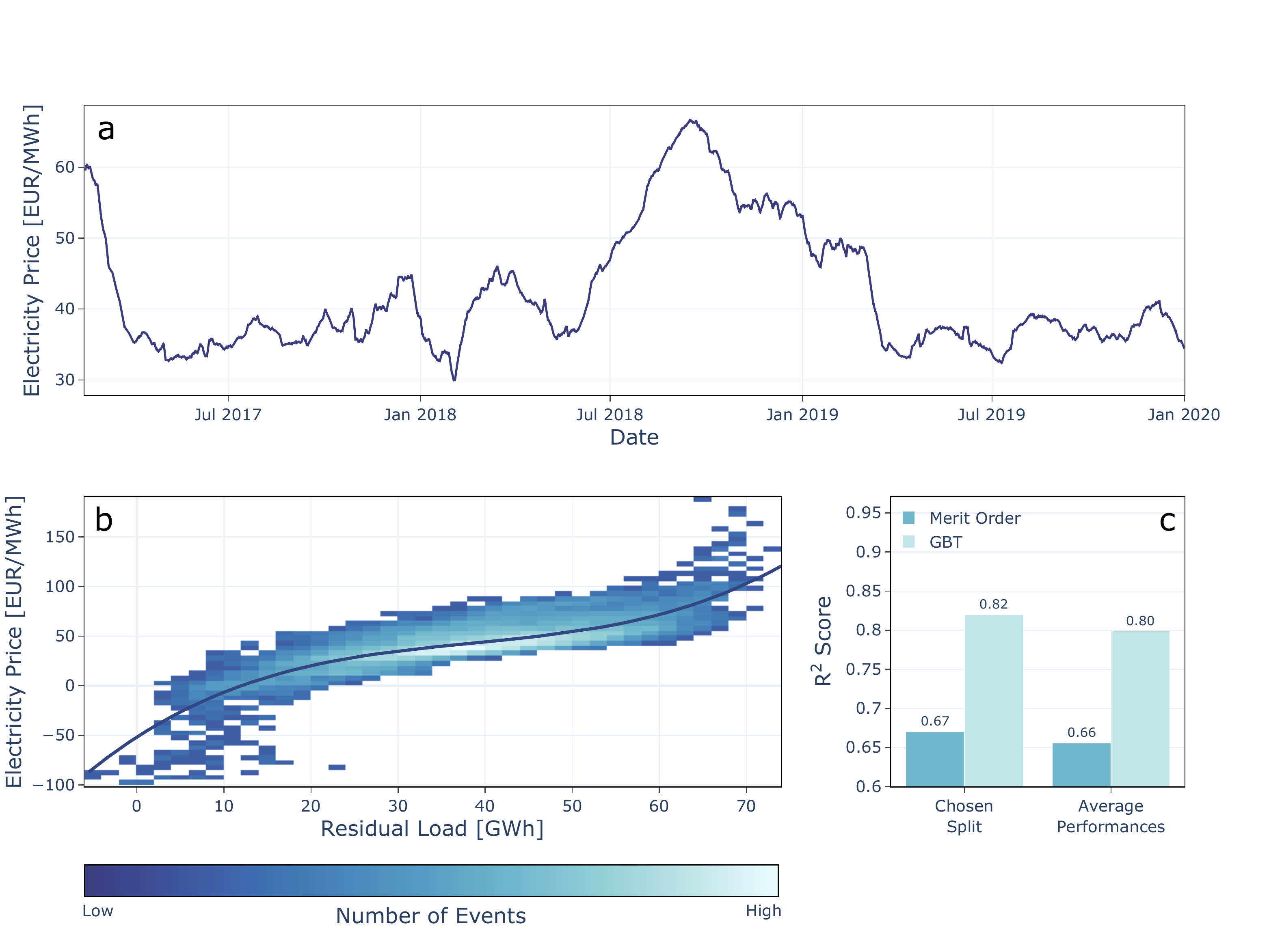}
    \caption{
    Explainable Machine Learning for day-ahead electricity prices.
    (a) Electricity price time series from the German day-ahead EPEX spot market from January 2017 to 2020.
    (b) In a single-feature benchmark model based on the merit order principle, prices are a function of the residual load, i.e. the difference of load and non-dispatchable renewable generation. The colormap shows a 2D histogram of the raw data, the line is a 3rd order polynomial fit.
    (c) Performance of the benchmark model and a gradient boosted tree (GBT) model, measured by the $R^2$ score on the test set.
    The GBT model outperforms the single-feature benchmark model and reveals more information about the electricity market. 
    }
    \label{fig:intro}
\end{figure*}

\section{Background and Literature}
\label{sec:background}
\subsection{The German electricity markets}
\label{sec:elec-market}

Electricity is traded on many different markets until it is delivered at a time $t_0$. Basic trading is done in the forward, day-ahead and intraday markets, while long-term non-public contracts between electricity producers and consumers via over-the-counter agreements are also possible. Here, We give a short overview of the German market structure focusing on day-ahead markets. Notably, Germany shares its bidding zone with Luxembourg and also shared it with Austria until 1st of October 2018. In the following, we will refer to this shared bidding zone as \textit{German markets} for the sake of simplicity.

To ensure that power generation matches the load, electricity is traded until a day before delivery in the day-ahead market or even minutes before delivery in the intraday market. This makes adaptation to sudden events like weather changes or unavailability of power plants possible and therefore reduces the need of expensive control reserves \cite{kruse_secondary_2022}. Since the intraday markets are smaller in volume and open after the day-ahead markets are closed, they are mainly driven by the day-ahead prices and react to short-term system changes like weather forecast error. This leaves the day-ahead markets as the main driver for general electricity prices reacting directly to availability of generation and load capacity.

Day-ahead trading is possible via different exchanges. In Germany the three operating exchanges are EPEX Spot SE, Nord Pool EMCO AS and EXAA AG \cite{acer_nemolist}. Trading is possible from 10:00 until 12:00 the day before delivery at $t_0$, where all exchanges are coupled with the pan European Single Day-Ahead Coupling (SDAC) to create one coupled market clearing prices (MCP) for all bidding zones \cite{Monopolkommission}. The MCP is calculated using an algorithm called EUPHEMIA, by using the bids and offers from all exchanges and the network constraints provided by the responsible TSOs.

In a simple and uncoupled market, electricity producers place their offer, which is a specific amount of electricity offered for a specific price for a given time period. Oppositely, electricity consumers place their bid, representing a specific amount of electricity they want to consume for a specific price. Once the order book closes, meaning no bids and offers can be placed anymore, the market clearing price is determined as the highest offer that finds a matching bid in each respective time window. The market clearing price is paid by every market participant \cite{han_complexity_2022}. Looking at the European markets, the SDAC enables cross-border trading, where offers and bids can be matched over different bidding zones, while also respecting network constraints.

Energy in the day-ahead markets is mainly traded for all 24 hours of the next day separately while there exist some smaller trading time frames and also specific blocks of important hours e.g. which represent base- and peak-load, depending on the operating exchange.

\subsection{The merit order principle and a single-feature price model}
\label{sec:merit-order}

Models based on the merit order principle provide a first approximation for the outcome of the day-ahead electricity market \cite{sensfuss2008merit}. We briefly review this approach, which will later serve as a benchmark model for more advanced machine learning models.

The electricity demand or load $L$ is mostly inelastic in the short term. Private consumers and small enterprises typically have fixed tariffs, hence the market price $p$ does not affect their short-term consumption behavior. Demand side management aims to increase the short-term flexibility and elasticity of demand, such that it can be adapted to availability of renewable power generation. However, the total amount of flexible loads is still limited. Hence, we assume that $L$ depends only on the time $t$, but not on the price $p$ for the time being.

In perfect competition, the demand is satisfied by generating units according to their marginal costs. All units with marginal costs below the market price $p$ can realize positive contribution margins and are thus ``on'', all others are ``off''. Hence one can obtain an approximate view of the market outcome by sorting all generating units according to their estimated marginal costs. Renewable power plants, in particular wind and solar power, have high investment costs but almost vanishing variable costs.
As marginal costs are dependent on the variable costs only, the marginal costs of these two renewable power sources are usually neglected.
Furthermore, renewable plants are prioritized in the German market according to national regulations (``Erneuerbare Energien Gesetz''). Hence, we can assume that the total renewable generation $G_{\rm ren}$ is independent of the market price $p$. However, renewable generation depends on the weather and thus varies strongly with time $t$. If the availability of the dispatchable power plants varies little in time, we can assume that the generation $G_{\rm dis}$ depends only on the price $p$. In the German market, nuclear and lignite power plants have the smallest marginal costs and thus contribute first. 
Notably, the marginal costs of individual power plants are not known exactly and must be estimated.

We can now formulate the condition of market equilibrium. Supply and demand match if
\begin{equation}
   L(t) = G_{\rm ren}(t) + G_{\rm dis}(p(t)) .
\end{equation}
Solving for the price yields
\begin{equation}
   p(t) = G_{\rm dis}^{-1} \left[ L(t) - G_{\rm ren}(t)  \right],
   \label{eq:price-merit}
\end{equation}
where $G_{\rm dis}^{-1}$ denotes the inverse function. That is, the market price is a function of the difference of load and renewable generation, which is commonly referred to as the residual load. 

Figure \ref{fig:intro}b shows the price in the German day-ahead market as a function of the residual load. Data has been collected for 3 years and is displayed as a 2D histogram. We observe that the assumption (\ref{eq:price-merit}) provides a reasonable approximation of the actual market behavior -- the price generally increases with the residual load. We fit a third-order polynomial to the data, which will serve as a benchmark model in the following. We find that the data scatters quite strongly around this fit, as various effects not taken into account in this approximate treatment.

\subsection{Previous work on electricity price prediction and modelling}

Understanding and forecasting electricity prices is an important research question in energy science. 
Therefore, day-ahead electricity prices have been studied from many different perspectives using a variety of methods and models \cite{lago_forecasting_2021,praktiknjo2016renewable}. Data based approaches range from autoregressive models, where predictions of future prices are computed from the historic time series alone \cite{contreras_arima_2003, conejo_day-ahead_2005}, to regression models including many different external features. Which features are important for electricity price predictions is still a matter of current research. For instance, the importance of weather forecasts has been highlighted in \cite{sgarlato2022role, goodarzi_impact_2019}.  Most recently, more complex machine learning models have been proposed for electricity price forecasting \cite{lago_forecasting_2021}, using many different approaches such as convolutional neural networks (CNNs) \cite{khan_short_2020} or recurrent neural networks (RNNs) \cite{li_day-ahead_2021,iwabuchi2022flexible}.

Many machine learning approaches are based on black-box models, which limits scientific insights \cite{roscher2020explainable, adadi_peeking_2018} and may induce security risks in critical sectors \cite{ahmad_artificial_2021, cremer_optimization-based_2019}. A promising alternative is provided by methods from eXplainable artificial intelligence (XAI), including inherently transparent models as well as post-hoc model explanations \cite{barredo_arrieta_explainable_2020}. The field of XAI has gained a strong interest in energy systems analysis in recent years \cite{machlev_explainable_2022}, in particular for applications power system operation and stability. For instance, XAI was used in transient stability assessment \cite{chen_xgboost-based_2019}, the identification of risks for frequency stability \cite{kruse_revealing_2021} or load and renewable generation forecast \cite{mitrentsis_interpretable_2022}. Furthermore, XAI has been used to analyse factors that determine the success of large power system infrastructure projects \cite{alova_machine-learning_2021}. Applications of XAI methods for electricity markets are still in its infancy. A recent study by Tschora et al \cite{tschora_electricity_2022} mainly focused on the identification of relevant features in forecasting models. 

Overall, machine learning methods have been almost exclusively used for electricity price forecasting, not for explanations. However, understanding which factors determine electricity prices is an important question in energy science, including for instance the impact of fossil fuel prices \cite{zakeri2022energy}.

In contrast to data-based methods, model-based approaches try to predict prices from fundamental economic considerations. There exist several mechanistic models that haven been developed to simulate electricity markets and explain the emerging electricity prices. For instance, agent-based models have been developed to simulate market processes using a variety of external features from many areas and explain the emerging electricity prices \cite{reeg_amiris_2012, nitsch_economic_2021}. Additionally, optimization models are routinely used to determine the dispatch and the market price \cite{wood2013power,stoft2002power}. Since simulation and optimization models are inherently transparent they provide an explanation of electricity prices from economic principles. However, they are limited to the mechanisms and interactions explicitly included by the modeler. XAI models can complement this approach by identifying key features, their dependencies and interactions, which in turn could inspire model extensions.

\section{Methods}
\label{sec:methods}

We develop an explainable machine learning model to understand German day-ahead electricity prices beyond the merit order effect introduced in Sec.~\ref{sec:merit-order}.

Since we focus on explaining our model, we will not make predictions in the sense of a forecast. Predictions are performed for modelling and analysing the electricity market.

For any further technical details about the methods the full project code is available on GitHub \cite{github}. This includes data preparation, hyperparameter optimization and model explanation.

\subsection{Data}

As our prediction target, we use the hourly day-ahead electricity prices for Germany, which we collect from the ENTSO-E transparency platform \cite{entso-e_transparency}. Since European day-ahead market prices are coupled via the SDAC explained in Sec.~\ref{sec:elec-market}, we get only one price for all exchanges.

It is important to note that Germany shares its bidding zone with Luxembourg and also shared it with Austria until 1st of October 2018 \cite{eex_vorbereitung_2018}. Throughout the article, \textit{German electricity prices} denote the price in the bidding zone of the given time period. Prices before and after the change of the bidding zones are joined together to create one continuous time series (see Fig.~\ref{fig:intro}a).

\begin{figure*}[tb]
    \centering
    \includegraphics[width=0.9\textwidth]{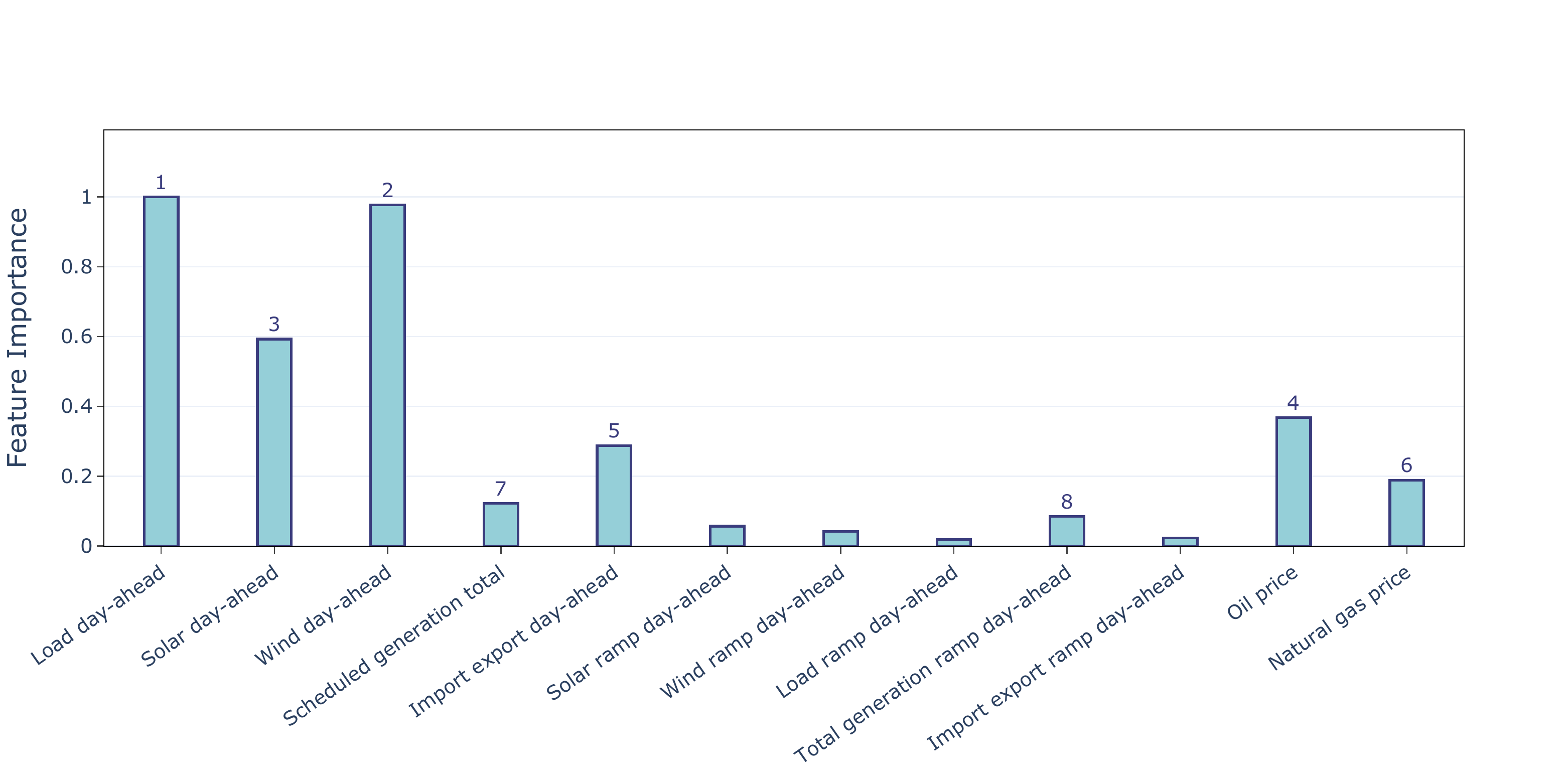}
    \caption{
    Feature importance in the GBT model for the day-ahead electricity prices.
    Feature importances are computed from SHapley Additive exPlanations and normalized to one (see text for details).
    Features contributing to the residual load are most important as expected, but fuel prices also rank high.
    }
    \label{fig:feat_imp}
\end{figure*}

As inputs for our prediction model we use power system features and fuel prices. Power system features are collected from the ENTSO-E transparency platform \cite{entso-e_transparency} and fuel prices are collected from ARIVA.DE AG \cite{ariva}. A full list of the features can be seen in Fig.~\ref{fig:feat_imp}.

Power system features include day-ahead forecasts of load, solar generation, wind generation, the day-ahead total generation and imports and exports. The features are aggregated for the four control areas of 50Hertz Transmission, Amprion, TenneT and TransnetBW. Wind generation is aggregated from wind on- and offshore generation. Total generation corresponds to the total scheduled generation in the day-ahead market. Import and export is aggregated from the cross border flows between Germany and the neighbouring bidding zones, where a positive (negative) value corresponds to more energy imports (exports). 
We also supplement the power system features with ramps for each feature, which are calculated using the formula $\mbox{ramp}(t)=f(t) - f(t-1)$ where $f(t)$ denotes the feature at a point $t$ in time.

Fuel prices include oil prices and natural gas prices. Because both features have a daily time resolution we create a linear interpolation to get an hourly time resolution matching the time sampling of the model.
Coal prices vary only very little during the considered time span.
We note that we exclude CO${}_2$ prices, although they affect electricity prices in the long-term. During the considered time period, CO${}_2$ prices have increased almost monotonically allowing the ML model to memorize the train set leading to immediate overfitting. In particular, we tested models including the CO${}_2$ price and found a high generalization error.

We use 3 years of data from the years 2017, 2018 and 2019 in order to get enough data for training and evaluation. Hourly data points with missing values for any feature are dropped to prevent fitting corrupted data.

\subsection{Model}

To model the German electricity price, we use Gradient Boosted Trees (GBTs) on our input data consisting of power system and fuel price features. GBTs offer complex non-linear models which we need in order to get more precise predictions of the electricity price than the benchmark model based on a common approximation of the merit order principle \cite{chen_xgboost_2016}. We use the LightGBM framework for our implementation in order to achieve a fast model training \cite{ke_lightgbm_2017}. 

While single decision or regression trees are interpretable by reporting their decision path, ensemble methods, such as GBTs, trade a higher performance for a harder-to-interpret model. Still, using methods such as SHAP enables us to get a detailed explanation of the GBTs which we explain in detail in Sec.~\ref{sec:shap}. 

For the training process, we split our data into a training (48\%), validation (32\%) and test (20\%) set. Since we focus on explaining our model instead of forecasting electricity prices as mentioned above, we shuffle our data before splitting. We use a weekly shuffle, where we only shuffle the dataset by weeks instead of hours before splitting. This gives us a more general model because we reduce the amount of similar data points in the training, validation and test set. We use the $L^2$ loss for the training process and the corresponding $R^2$ score for evaluating the performance of the models. 

During training the validation set is split into 4 equally large sets to avoid overfitting of the validation set. The 4 validation sets are used for evaluation of the performance after each training epoch, where the training is stopped if the performance of one of the validation sets is not improving for a specific number of epochs.

We use a random search to find the best hyperparameters where we evaluate the performance of the fully trained models on the unseen test set. We choose the model with best performance on the test set. 

For specific analysis tasks we need to ensure the consistency of our models. We achieve this by analysing the 10 best models of our random search for 10 different weekly random splits.

\begin{figure*}[tb]
    \centering
    \includegraphics[width=0.9\textwidth]{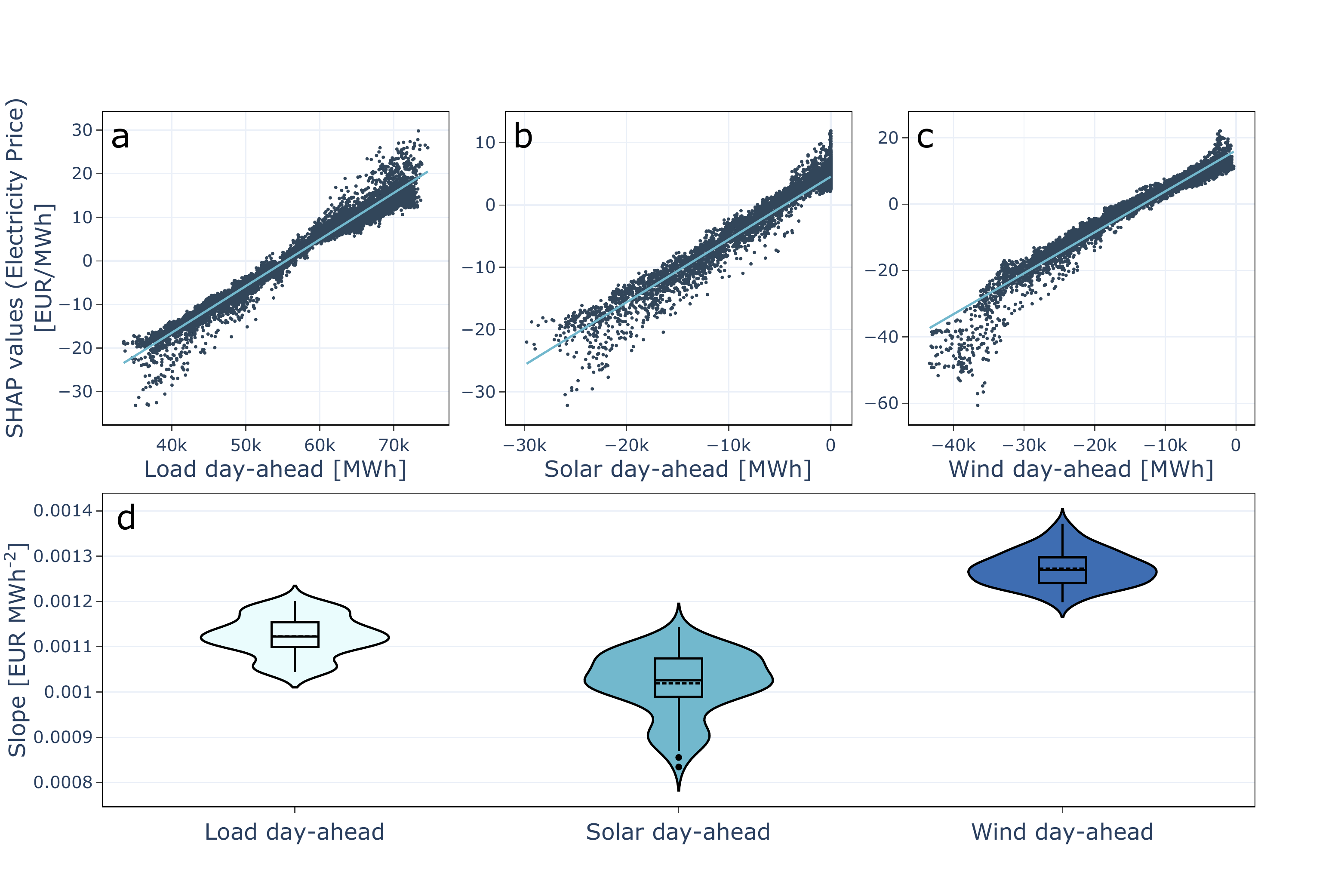}
    \caption{
        Impact of the residual load features in the GBT model. (a)-(c)  Dependency plots for the residual load features load, solar generation and wind generation (measured by mean absolute SHAP values). The light blue line is a linear fit. 
        (d) Slope of the linear fits in the dependency plots. Violin plots shows the results for the ten best models for ten different data splits. 
        The residual load features have slightly different linear relation to the electricity price, which is not captured by the benchmark model based on the merit order principle.
    }
    \label{fig:res_load_depend}
\end{figure*}

\begin{figure*}[tb]
    \centering
    \includegraphics[width=0.9\textwidth]{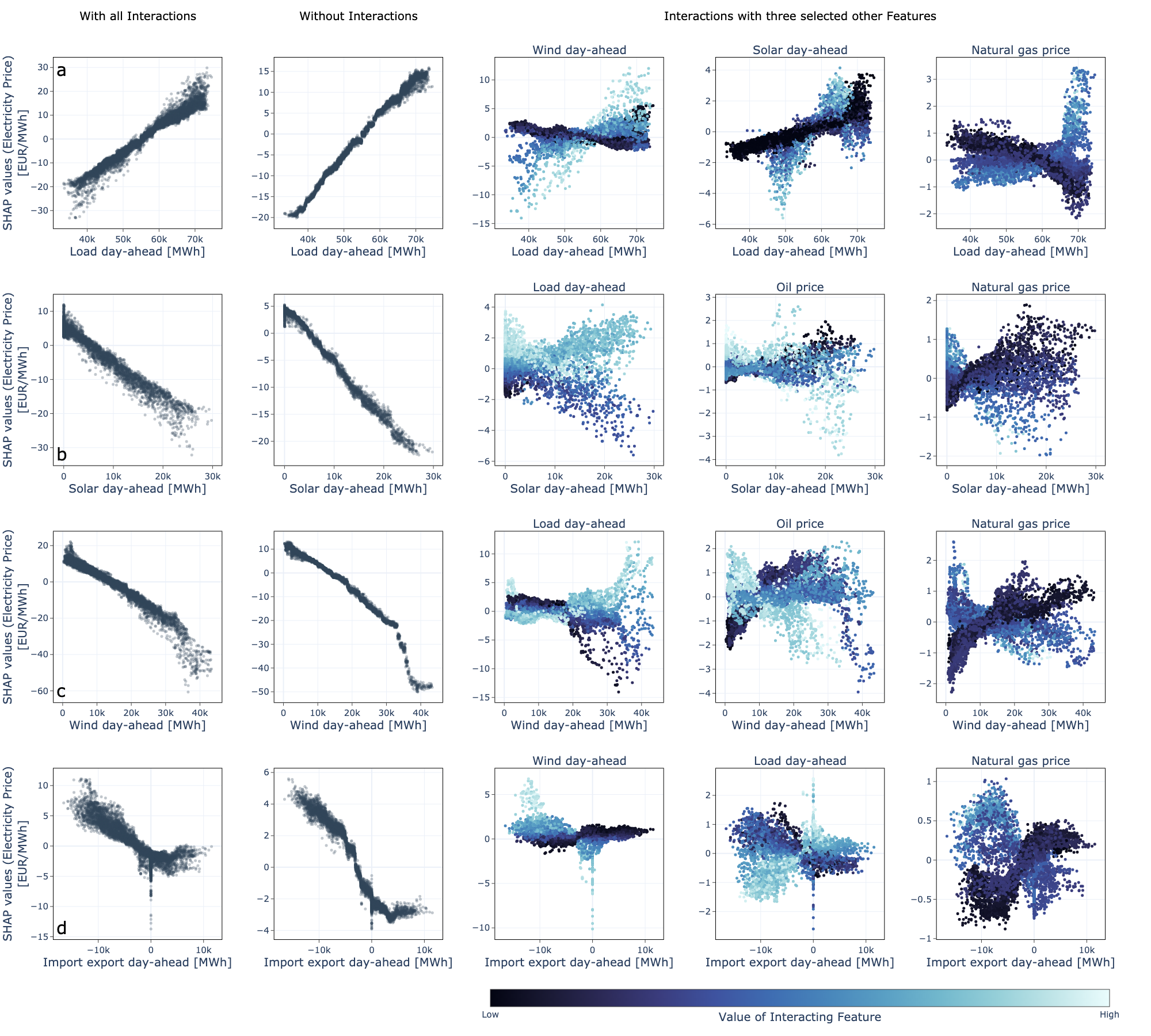}
    \caption{
        Dependencies and interactions for (a) the load, (b) solar generation, (c) wind generation and (d) cumulative import or export, respectively.
        The first column shows the full SHAP dependency plot, the second column the SHAP dependency plot without any interactions. The third to fifth column show the SHAP interactions of three selected features, where the color indicates the value of the interacting feature. 
        Fuel prices play an important role when interacting with other features in the GBT model. Further details are given in the text.
    }
    \label{fig:res_load_int}
\end{figure*}

\begin{figure*}[tb]
    \centering
    \includegraphics[width=0.95\textwidth]{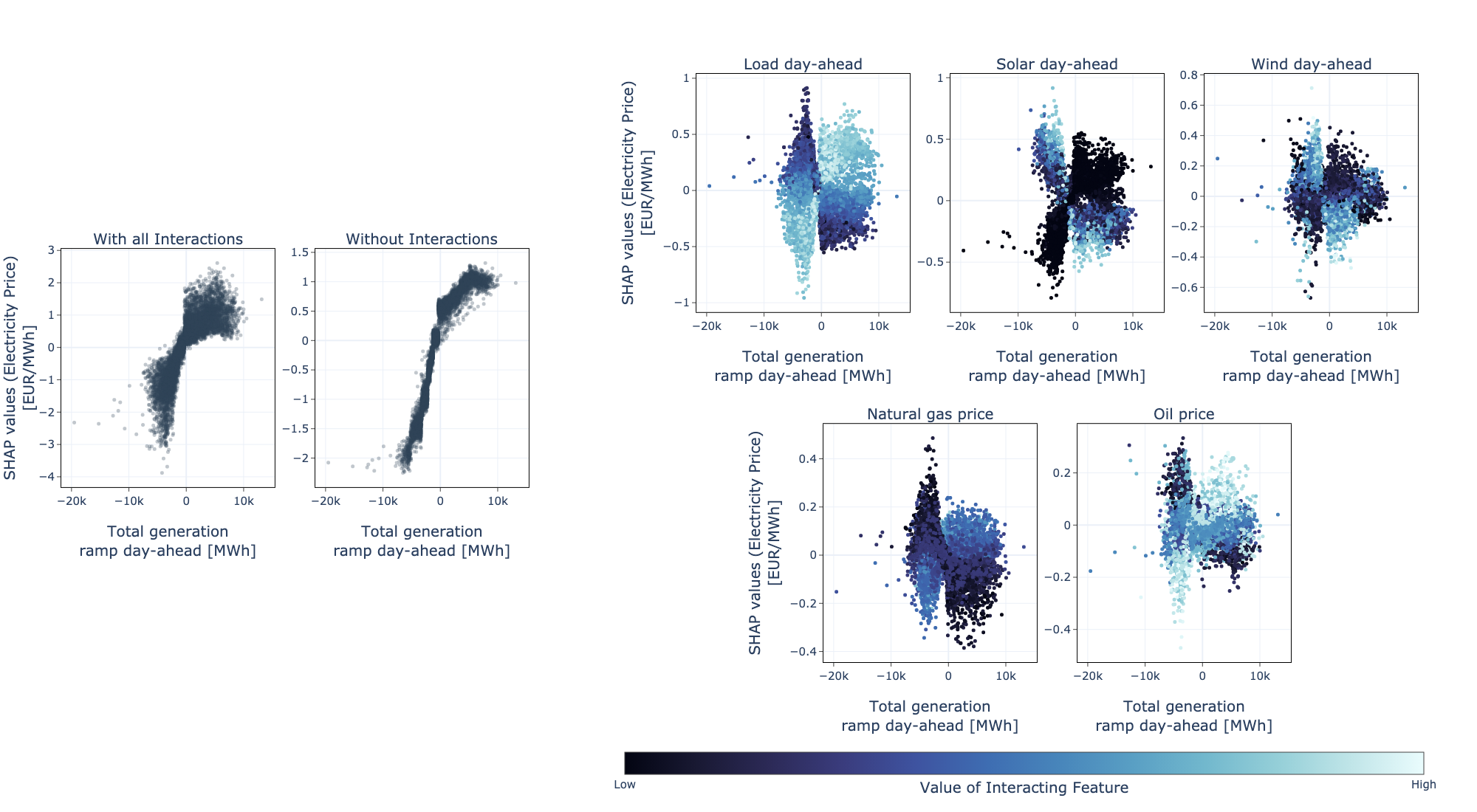}
    \caption{
        Effect of generation ramps and limited generation flexibility on electricity prices.
        The first column shows the full SHAP dependency plot of the total generation ramp, the second column the SHAP dependency plot without any interactions.
        Further panels show the SHAP feature interactions for five selected interacting features. The color indicates the value of the interacting feature. 
        Ramps affect/alter the electricity price prediction by up to 10\%, depending on factors as load and renewable generation, but also on fuel prices. 
    }
    \label{fig:ramp_int}
\end{figure*}

\begin{figure*}[tb]
    \centering
    \includegraphics[width=0.9\textwidth]{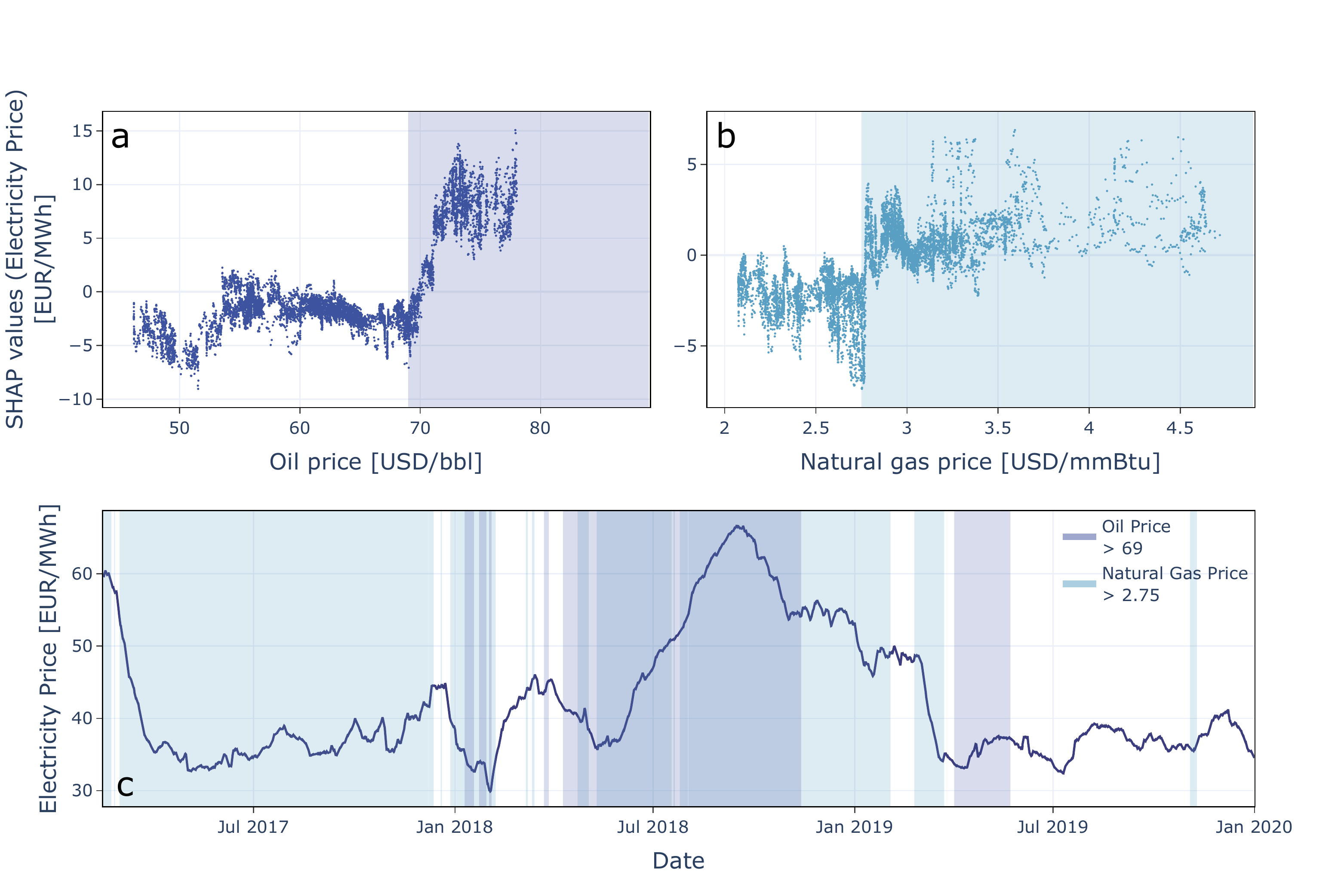}
    \caption{
        Impact of oil and gas prices on the electricity price predictions. 
        (a)-(b) SHAP dependency plot of the oil price (gas price), where the colored area marks a change for the dependency in the GBT model. (c) Electricity price time series from the German day-ahead EPEX spot market from January 2017 to 2020. Dark (light) blue areas mark the time periods with oil (gas) prices above the dependency threshold. 
        We observe a clear change in the dependency on fuel prices in the model, but it cannot be asserted whether this is a causal relation. 
    }
    \label{fig:oil_gas_depend}
\end{figure*}

\subsection{Model explanation}
\label{sec:shap}

The model is interpreted using SHapely Additive exPlanation (SHAP) values after training \cite{lundberg_local_2020, lundberg_unified_2017}. SHAP values quantify the positive or negative impact of each feature relative to the overall base value. In our case, SHAP values show which features lead to higher or lower price predictions. They avoid inconsistencies present in other feature attribution methods and fulfill desirable properties for explainable machine learning \cite{lundberg_consistent_2019, lundberg_unified_2017}.

For example \textit{local accuracy} guarantees that SHAP values sum up to the model prediction, meaning all feature contributions plus a base value \cite{lundberg_consistent_2019}. Overall for the $n$ feature values $x_1,\dots,x_n$ the output of the model $f(x_1,\dots,x_n)$ can be written as 
$$f(x_1,\dots,x_n) = \phi_0 + \sum_{j=1}^n \phi_j(x_1,\dots,x_n),$$
where $\phi_0$ corresponds to the base value of the model predictions and $\{\phi_j\}_{j=1}^n$ are the SHAP values of the corresponding features.

From the \textit{local} SHAP values we quantify a \textit{global} feature importance by averaging over all SHAP values for each feature and normalizing them by the highest value (see Fig.~\ref{fig:feat_imp}). We also use SHAP dependency plots (see Fig.~\ref{fig:res_load_depend}a-c), which give detailed insights into the feature contributions and utilize SHAP interaction plots to get insights about feature interactions inside the model (see Fig.~\ref{fig:res_load_int}~and~\ref{fig:ramp_int}).

\section{Results}
\label{sec:results}

The developed machine learning model is capable of predicting day-ahead electricity prices with an average performance of $R^2 = 0.8$ (Fig.~\ref{fig:intro}c). Roughly speaking, the model explains 80 \% of the variability of price time series. The performance is substantially better than for the benchmark model based on a common approximation of the merit order principle reaching only $R^2 = 0.66$. We conclude that the machine learning model captures several market effects which are neglected in the single-feature benchmark model described in Sec.~\ref{sec:merit-order}. We will now discuss these effects in detail, interpreting the machine learning model with the SHAP framework.

\subsection{Features affecting the electricity prices}
\label{sec:feat_imp}

The developed machine learning model takes into account a variety of different features beyond the residual load. The SHAP values provide a consistent measure of the feature importance and thus reveal which factors have the strongest influences on the prices. Cumulative feature importance are shown in Fig.~\ref{fig:feat_imp}.

As expected, the main driver of the electricity prices is given by the residual load. More precisely, the three residual load features (load, wind and solar generation) are also the most important features in the machine learning model. The dependency of these features will be discussed in more detail in Sec.~\ref{sec:load-wind-solar}.

Fuel prices rank at position 4 and 6, with oil prices being more important than gas prices. This dependency is not surprising as fuel prices directly affect the variable costs of the respective power plants. However, the precise interpretation of this finding is less clear and will be further discussed in Sec.~\ref{sec:fuel-prices}.

Prices are obviously related to cross-border trading. The import-export balance is the fifth most important feature and will be discussed in detail in Sec.~\ref{sec:imports}. The total generation and its ramp rank at position 7 and 8. The generation ramp is particularly interesting as it reveals the influence of previous time steps, see Sec.~\ref{sec:ramps} for details.

\subsection{The role of wind, solar and load}
\label{sec:load-wind-solar}

The residual load features, i.e. load, solar generation and wind generation, are the most important features for the machine learning model (Fig.~\ref{fig:feat_imp}), in agreement with the benchmark model based on the merit order principle explained in Sec.~\ref{sec:merit-order}. We take a more detailed look at the contribution of the residual load features by analyzing the corresponding partial dependency plots and interaction plots.

The benchmark model assumes that the price depends only on the residual load, hence the three features enter in an equal way up to a sign. Analyzing the respective partial dependency plots in Fig.~\ref{fig:res_load_depend}a-c, we observe a similar dependency as expected, but also some subtle differences. 
For the detailed analysis of the differences and the observed scattering we simplify the comparison of the three features by multiplying the renewable generations by $-1$.

The dependency on load, wind and solar generation is approximately linear (Fig.~\ref{fig:res_load_depend}a-c), hence we use a linear fit for a quantitative analysis. We create linear fits for the partial dependency plots for the 10 best models after random search of 10 different random weekly shuffled splits. Figure~\ref{fig:res_load_depend}d shows the slope of the linear fits on the dependency plots for the 100 different models as violin plots. The different models seem to be consistent with their dependencies since the violin plots show a clear distribution around the mean value of the slope. Only the violin plot for solar generation shows some outliers.

The benchmark model based on the merit order principle assumes an equal contribution of all residual load features, but the machine learning model reveals some subtle differences. 
The slopes of the dependency on load and solar are rather similar, with solar being slightly smaller. In contrast, the slope of the dependency on wind is notably larger. 
The smaller influence of solar generation may be due to the fact that solar generation is more distributed in the German power grid, with a large fraction installed directly at the consumers. This could lead to solar generation acting as a negative load in the power grid, which would naturally cause load and solar generation to have a similar impact on the electricity price. In contrast, the generation of wind is more concentrated in the north of Germany, especially in the case of off-shore wind generation. This could lead to wind generation acting more like other power plants in the energy system, making the presence or absence of wind generation more important for the electricity price.

A further reason for the different role of wind and solar may lie in their respective  market rules. While small scale PV installations typically rely on fixed feed-in tariffs, wind turbines are incentivized to sell their power according to the wholesale electricity market prices (`Direktvermarktung', see \cite{2014eeg}).

The small scattering visible in the dependency plots can be explained by feature interactions. Neglecting all interactions in the dependency plot in Fig.~\ref{fig:res_load_int}a-c second column, scattering becomes smaller and the dependency of the residual load features is even clearer. The dependency on load and solar generation is approximately linear, while the dependency on wind shows a clear non-linearity for large in-feeds above 30 GW. In this case, flexible power plants such as natural gas or oil plants have already left the market. Then, market equilibrium requires either an  increase in exports, an increase of the load or a reduction of generation from mostly inflexible power plants such as lignite or nuclear. It appears plausible that these three mechanisms are comparatively inelastic such that the price decreases rapidly.

Looking at three of the most important interacting features, we can attribute most of the scattering present in the normal partial dependency plots. 

Load has its strongest interactions with the renewable generation wind and solar. Both interactions are similar and enhance the dependency on load. This is reasonable since these features combined serve as a good approximation for the residual load, which is again already a good predictor for the electricity price. We can also see a strong interaction with the gas price, especially for high gas prices and high load. This is because more gas power plants are active for higher load, which then leads to higher electricity prices if gas prices are also high. 

Wind and solar generation have a similar interaction structure, in particular there is a strong interaction with the load but also with fuel prices. For the load, interactions are particularly strong for high renewable generation. If the load is small, we recover the situation discussed above for the case of high wind generation, where market equilibrium requires the adaption of comparatively inelastic participants and thus entails strong prices signals which may even include negative prices. This effect is largely compensated if the load is also high, leading to a strong increase of the price. 

We also see strong interaction with the fuel prices, both amplifying the dependency for low wind or solar generation and reducing it for high wind or solar generation. These interactions originate from the fact that more fuel-dependent power plants are active if renewable generation is low and therefore the electricity price is more dependent on fuel prices in this case. 

Summarizing, the residual load features have an overall strong interaction with fuel prices, mostly due to more dispatchable generation being active for specific values of the residual load features. Fuel prices not being included in the single-feature benchmark model could explain its lower performance compared to the GBT model.

\subsection{Prices and cross-border trading}
\label{sec:imports}

Electricity prices and trades are intimately related. In the machine learning model, the import-export balance ranks at fifth place in terms of the feature importance (Fig.~\ref{fig:feat_imp}). Different mechanisms can contribute to this dependency. On the one hand, high prices foster imports from other countries. On the other hand, imports provide a further source of electricity and should thus lead to lower prices. These two interactions can be interpreted as opposite causal relations, where either the price or the import-export balance is acting as the driver. 

However, we stress that a causal interpretation is not that straightforward. The prices in different bidding zones and the cross-border trades are not determined sequentially, but simultaneously via the EUPHEMIA algorithm \cite{Monopolkommission}, see also
Sec.~\ref{sec:elec-market}.   

The SHAP dependency plot reveals a negative correlation of the import balance and the day-ahead price (Fig.~\ref{fig:res_load_int}d). That is, imports are typically related to lower prices, while exports are related to higher prices. To understand this correlation, we consider one specific market situation. Assume that there is a high wind power generation in Germany. Typically, there are many low price offers in the German bidding zone, leading to a low market clearing price. However, if here is a strong demand in a neighboring country, additional offers will be accepted for exports, leading to an increase of the market clearing price. Hence, it appears as if the exports drive the market price, but in fact both are driven by a common cause: the total supply and demand in the two neighboring countries combined.

The strongest feature interactions are found for wind power generation and load. The interaction is opposite, which is comprehensible because the two feature enter the residual load with opposite sign. We find that an increase of the residual load reduces the observed dependency, while a decrease of the residual load increases it.

Since SHAP values reveal only correlations of features and targets, it is difficult to reach a comprehensive causal interpretation. Still, based on our results, we formulate the following hypothesis. A high demand from a neighboring country generally leads to exports and to an increase of prices. But if the domestic demand (the residual load) is also high, there are no cheap offers left in the order book that would allow for exports. Hence, the dependency of exports and prices diminishes. 

Notably, there is large scatter to lower prices in the case of a vanishing import export balance. This might be due to a temporary reduction in the transmission capacity preventing exports, or corrupted data (see \cite{hirth2018entso} for a discussion of the data quality of the ENTSO-E transparency platform).

Finally, we also see an interesting interaction of the import-export balance with gas prices. Lower gas prices tend to reduce the overall contribution of import-export while higher prices amplify this contribution. This could indicate that at low gas prices, local gas generation reduces the dependency of a country to exchange power with neighboring countries and hence the price is influenced less by its imports and exports. In the opposite case of high gas prices, countries will be more willing to exchange energy and the effect of cross-border flows on prices increases.

\subsection{Impact of ramps} 
\label{sec:ramps}

The machine learning model reveals a weak dependency of the price and the power generation ramps (Fig.~\ref{fig:feat_imp}). Hence, the market outcome in a certain hour is affected by previous hours. The partial dependency plot (Fig.~\ref{fig:ramp_int} left) indicates a positive correlation of the price and the total generation ramp. Hence, prices tend to be higher if generation is ramped up and lower if generation is ramped down.

This finding can be attributed to a limited flexibility of conventional  power plants, in particular nuclear and lignite plants~\cite{Bergh2015cycling}. First, technical limits exist for the ramping speed and the minimum generation in partial load. Second, ramping and cycling induce additional costs, for instance due to a wear and tear of the power plant. Hence, there is an incentive to limit generation ramps which can affect the bidding on the market. In case of a decreasing total generation, operators may bid at a lower price to remain in the market and avoid ramping downwards. Similarly, in case of an increasing total generation, operators may bid at a higher price. As a consequence, the price increases with the total generation ramp.

The role of generation ramps depends on several other factors. The SHAP dependency plot is strongly scattered which can be attributed to the presence of feature interaction.
A high value of the load amplifies the impact of generation ramps. If the load is high, more conventional generation is needed in general, such that ramping limits and costs are more important. Vice versa, high values of 
wind and solar generation mitigate the impact of generation ramps as less conventional plants are needed.
Furthermore, high oil and gas prices amplify the impact of generation ramps, too. Oil and gas power plants typically have a higher flexibility than nuclear or coal power plants. Higher fuel price penalize these plants, such that nuclear or coal power plants may have to contribute more strongly to the ramping process.

\subsection{Impact of fuel prices: Correlation or causality?}
\label{sec:fuel-prices}

Looking at the feature importance in Fig.~\ref{fig:feat_imp} we note that the machine learning model is critically dependent on fuel prices. The oil price is the 4th most import feature, i.e.~the most important feature after the residual load features, while the gas price is the 6th most important. We analyze the dependency on oil and gas prices in detail.

The dependency plots for the fuel prices in Fig.~\ref{fig:oil_gas_depend}a-b are strongly nonlinear, with an almost step-like behaviour.
The dependency of the electricity price on the oil price is approximately constant below a threshold of $69\,\mbox{[USD/bbl]}$. Above this threshold, the dependency increases until it saturates.
While the change in dependency is almost linear for oil prices, the dependency for gas prices shows a step-like behaviour, with a threshold at $2.75\,\mbox{[USD/mmBTu]}$, albeit with a stronger scattering. A causal interpretation is comprehensible as an increase in fuel prices leads to an increase in the operational costs of the respective power plants and thus to offers at higher prices.

For further analysis, we focus on the time series of the electricity price in Fig.~\ref{fig:oil_gas_depend}c, highlighting the time periods with oil and gas prices above the dependency threshold respectively. High oil prices seem to be correlated with high electricity prices, especially for the maximum of electricity prices at the end of 2018. In contrast, in the middle of 2019, electricity prices stay low while oil prices are above the threshold. For gas, the dependency is even less clear. Lower prices seem to be a proxy for low electricity prices in 2019. Meanwhile, high gas prices do not display a clear correlation with overall electricity prices in the time series.
We further note that a delayed relation could also be possible, if power plants purchase fuel well before the usage.

In general, it is difficult to pinpoint the relation of fuel prices to the electricity price beyond statistical correlations. Although we find a strong change in the dependency on the fuel prices in the machine learning model, it is still possible that the machine learning model is using the fuel prices to remember specific time periods where electricity prices are higher or lower than expected from the other features. Nevertheless, we find reasonably strong interactions of the residual load features with fuel prices, as discussed in  Sec.~\ref{sec:load-wind-solar}. This points to a causal interaction, but a confounding effect is also possible. Overall, fuel prices are correlated with electricity prices which could be one of the reasons the machine learning model outperforms the benchmark model.

\section{Discussion and Conclusion}
\label{sec:discussion}

Summarizing, we have developed a machine learning model based on gradient boosted trees and demonstrated how it accurately estimates electricity prices, outperforming a single-feature benchmark model based on a common approximation of the merit order principle. Using SHAP to interpret our black-box model, we obtained deeper insights into the characteristics of the day-ahead market. SHAP values quantify how the price depends on the input features and thus reveals drivers beyond the benchmark model. Our analysis confirmed that high load leads to high prices, while large shares of wind or solar generation reduces prices. Furthermore, the model quantified the role of fuel prices, imports and exports, as well as load and generation ramps.

We saw that the SHAP analysis of our model is limited when it comes to a deeper causal interpretation of feature impacts. Nevertheless, the SHAP values provide detailed insights into the working of the model by revealing how and which features are mostly used and by quantifying dependencies and interactions.
Only by including domain knowledge of market mechanisms and power systems we can hypothesise to causal relations. Estimating causal models directly from data is in principle possible, e.g. using Causal SHAP \cite{heskes2020causal} or causal representation learning \cite{scholkopf2021toward} but requires more explicit assumptions about the underlying causal structure than we wanted to employ in this first exploratory study.

Concluding, we have demonstrated the usefulness of XAI models to analyse electricity price dynamics in the German market. Model-based approaches can benefit from the insights gained by the SHAP analysis. For instance, our results suggest slightly different roles of wind and solar power, while they enter the residual load equally. Furthermore, the SHAP analysis quantifies the role of generation ramps, which are subject to strong feature interactions.

There remain many open questions and starting points for further research. It would be interesting to investigate how XAI price models differ between electricity markets in different countries. Similarly, XAI models may also be used to compare markets at different times to quantify changes. For instance, the phase-out of nuclear power in Germany or changes of the regulatory framework should impact the dependencies of prices and features.
Furthermore, XAI methods may also be used to analyze the impact of exceptional events such as the energy crisis after the Russian invasion of Ukraine once sufficient data is available.

\section*{Acknowledgments}

We thank Sebastian Pütz, Johannes Kruse and Maurizio Titz for valuable discussions.
We gratefully acknowledge support from the Helmholtz Association via the grant no.~VH-NG-1727 to B.S. and the grant \textit{Uncertainty Quantification -- From Data to Reliable Knowledge (UQ)} no.~ZT-I-0029 to D.W. 
A.P. acknowledges funding under the Excellence Strategy of the Federal Government and the Länder (grant no.~ECU M2 ERS Funds OPSF689).


\end{document}